%% file: main.tex
\documentclass[letterpaper, 10 pt, conference]{ieeeconf}  

\IEEEoverridecommandlockouts                              

\usepackage{graphics} 
\usepackage{epsfig} 
\usepackage{mathptmx} 
\usepackage{times} 
\usepackage{amsmath} 
\usepackage{amssymb}  
\usepackage{color,soul}
\usepackage{multirow}
\usepackage{color}
\usepackage{graphicx}
\usepackage{subcaption}
\usepackage{subcaption}

\input{tex/math_defs.tex}

\newcommand{\add}[1]{\textcolor{black}{#1}}

\title{\LARGE \bf
BTEL: A Binary Tree Encoding Approach for Visual Localization
}

\author{Huu Le$^\dagger$, Tuan Hoang$^*$, and Michael Milford$^\dagger$
\thanks{$^\dagger$School of Electrical Engineering and Computer Science, Queensland University of Technology, Brisbane, Australia. $^*$Singapore University of Technology and Design. This work was supported by an Asian Office of Aerospace Research and Development Grant FA2386-16-1-4027 and an ARC Future Fellowship FT140101229 to MM. Email: huulem@outlook.com; michael.milford@qut.edu.au   }
}

\begin{document}

\maketitle
\thispagestyle{empty}
\pagestyle{empty}

\begin{abstract}        
        Visual localization algorithms have achieved significant improvements in performance thanks to recent advances in camera technology and vision-based techniques. However, there remains one critical caveat: all current approaches that are based on image retrieval currently scale at best linearly with the size of the environment with respect to both storage, and consequentially in most approaches, query time. This limitation severely curtails the capability of autonomous systems in a wide range of compute, power, storage, size, weight or cost constrained applications such as drones. In this work, we present a novel binary tree encoding approach for visual localization which can serve as an alternative for existing quantization and indexing techniques. The proposed tree structure allows us to derive a compressed training scheme that achieves \emph{sub-linearity} in both required storage and inference time. The encoding memory can be easily configured to satisfy different storage constraints. Moreover, our approach is amenable to an optional sequence filtering mechanism to further improve the localization results, while maintaining the same amount of storage. Our system is entirely agnostic to the front-end descriptors, allowing it to be used on top of recent state-of-the-art image representations. Experimental results show that the proposed method significantly outperforms state-of-the-art approaches under limited storage constraints.

\end{abstract}

\input{tex/introduction}
\input{tex/related_work}
\input{tex/approach}

\input{tex/results}
\input{tex/conclusion}

\bibliographystyle{IEEEtran}
\bibliography{main}

\end{document}

%% file: tex/math_defs.tex
\newcommand{\cD}{\mathcal{D}}

\newcommand{\bX}{\mathbf{X}}

\newcommand{\bS}{\mathbf{S}}

\newcommand{\bx}{\mathbf{x}}

\newcommand{\bp}{\mathbf{p}}
\newcommand{\bq}{\mathbf{q}}

\newcommand{\bw}{\mathbf{w}}

\newcommand{\bbR}{\mathbb{R}}

%% file: tex/introduction.tex
\section{Introduction}
\label{sec:introduction}

Visual place recognition plays a crucial role in many computer vision and robotics applications, as it underpins a wide variety of fundamental problems, including self-localization~\cite{sattler2018benchmarking}, loop-closure detection, large-scale structure from motion. Based on the visual information obtained from an RGB(-D) image captured by a camera (or a depth sensor), the goal is to identify the camera location within a pre-defined map.  This research topic has gained much attention recently, where several methods have been introduced that can achieve impressive localization results, even for challenging datasets containing large amount of images and undergoing extreme changes in visual conditions~\cite{lowry2016visual}.

In contrast to the structure-based approaches~\cite{sattler2018benchmarking}, which involve building 3D point clouds that represent the map and conducting 2D-3D matching during the query stage to search for the camera positions and orientations, retrieval-based techniques~\cite{milford2012seqslam} make use of visual similarity between image descriptors to assist the localization process. Methods based on image retrieval usually cast visual place recognition as a special instance of nearest neighbor search, i.e., the query image is compared against all instances in the database and its nearest neighbors is used to infer the query location.  In recent years, the use of visual similarity search algorithms has proven to be of significant impact to a large number of works on visual localization for robotics navigation~\cite{se2002mobile} and autonomous driving~\cite{lowry2016visual}. \add{Retrieval-based approaches are also more feasible in a large range of application domains with limited access to data from Global Navigation Satellite Systems (GNSS) and Visual Inertial Odometry (VIO) such as underground mining, tunnels and indoor localization}. In addition to the traditional methods for visual place recognition such as SeqSLAM~\cite{milford2012seqslam} or FAB-MAP~\cite{cummins2008fab}, the main contributing factors to the recent success of visual similarity search are arguably the advances in learning image representations~\cite{netvlad}, the strong developments in state-of-the-art quantization techniques and novel indexing methods~\cite{jegou2011product} that allow querying in large-scale databases.

\begin{figure}[t]
    \centering
    \begin{subfigure}[b]{0.45\textwidth}
        \centering
        \includegraphics[width=0.95\textwidth]{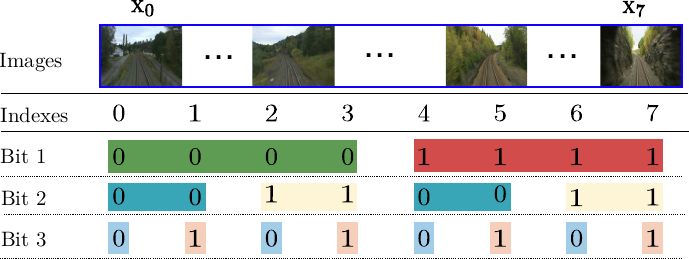}
        \caption{Image indices and their binary representations (for better visualization, two child nodes having the same parent are coded with different colors).}
        \label{fig:binary_representation}
    \end{subfigure}
    \begin{subfigure}[b]{0.45\textwidth}
            \centering
           \includegraphics[width=0.95\textwidth]{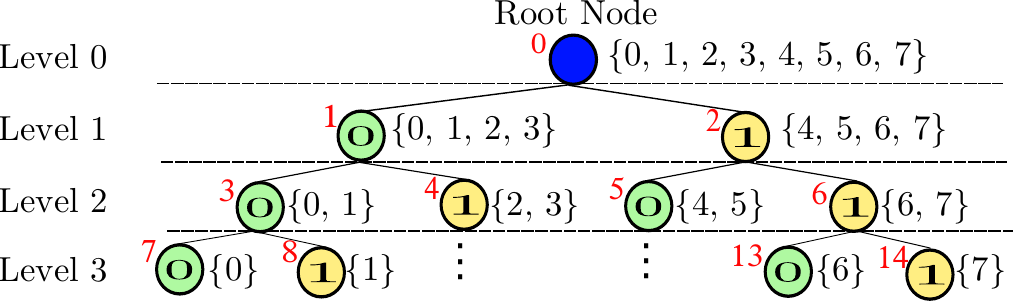}
           \caption{Tree Encoding corresponding to Fig.~\ref{fig:binary_representation}. The red numbers indicate the indexes of nodes in the tree.}
           \label{fig:tree}
           
    \end{subfigure}

    \caption{\add{Illustration of Binary Tree Encoding (BTEL) for an example dataset containing $8$ training images.} }
    \label{fig:tree_encoding}
    \vspace{-0.5cm}
\end{figure}

Although the aforementioned quantization and indexing techniques have substantially reduced the overall storage footprint and query time, their required memory to store the compressed data must increase linearly with the database size (i.e., number of training images). More specifically, regardless of the size of the compressed code vector, one code per training sample must be stored in the database. Therefore, when the number of training images in the database grows, the storage must also grow at least linearly. In addition, existing quantization approaches need large code-books in order to achieve satisfactory retrieval results, which in turn increases the total amount of storage required.


Recently, a sub-linear encoding algorithm was proposed in~\cite{yu2018rhythmic}, which makes use of the cyclic patterns to encode the scenes for the task of visual localization. However, this rhythmic representation approach can only work with applications where the scenes undergo small changes, which makes it impractical for large-scale datasets in varying visual conditions (e.g., day/night, weather or seasonal changes). Moreover, the requirement that the number of cyclic patterns must be co-prime causes much difficulty in choosing the parameters and leads to unnecessary storage being wasted during the training process. These weaknesses of~\cite{yu2018rhythmic} are also thoroughly addressed by our novel encoding approach.

Due to the growing demand of deploying visual search algorithms to mobile phones and robotic systems with limited storage capacity, there has been strong desire for storage-efficient encoding algorithms. For several applications such as long-term localization~\cite{krajnik2017fremen}, the continuous stream of data from sensory devices may result in an exponential growth of training storage. The development of storage-efficient encoding algorithms, especially algorithms that scale sub-linearly with the training data, would aid in the applicability of state-of-the-art visual localization techniques to real-world environments. \add{More generally and regardless of the absolute hardware constraints, sub-linear storage growth enables either improved scaling in operational envelope, an increase in the sophistication and complexity of information stored per place or a reduction in the computational and storage requirements in comparison to traditional methods.}

In this work, we address the aforementioned weaknesses of the existing encoding methods by introducing a new tree encoding approach that combines both quantization and indexing into one unified framework.   The main contributions of our work can be summarized as follows:
\begin{itemize}
    \item We propose a novel binary tree encoding (BTEL) method to \emph{directly learn} the location from the image descriptors. Unlike existing techniques that store one code vector per training sample, our method only stores the training parameters, resulting in a \emph{significant reduction} in storage footprint, while the performance is comparable to state-of-the-art techniques~\cite{gong2013iterative,jegou2011product}.
    \item We exploit the redundancy of the visual information encoded in image descriptors to derive a novel feature selection technique that significantly reduces the dimensionality of the feature vectors \add{while maximizing the separability of the data.} This approach yields a drastic reduction in memory footprint, training and querying times.
\end{itemize}
\add{Note that although our method utilizes a tree structure, it is fundamentally different from conventional tree-based methods for place recognition such as vocabulary tree~\cite{nister2006scalable} or traditional decision tree classifiers, as they require the number of nodes to scale at least linearly with the amount of training data}. With our proposed method, the encoding storage can be easily adjusted to satisfy the budget requirements in different applications, which makes it suitable for the deployment to devices with different storage and computational capabilities. Our method is able to achieve competitive localization results compared to state-of-the-art approaches under the storage constraint of 0.5MB. Moreover, our method is agnostic to the type of input features, allowing state-of-the-art image descriptors such as NetVLAD~\cite{netvlad}, DenseVLAD~\cite{torii201524} to be used. 
\add{Compared to~\cite{yu2018rhythmic}, which is the first method that achieves sub-linear storage growth for visual place recognition, our work proposes a more advanced encoding mechanism that achieves better  performance in terms of accuracy, scalability, and robustness to environmental changes.}

%% file: tex/related_work.tex
\section{Related Work}
\label{sec:related_work}
 \add{Many retrieval-based algorithms have been developed in the literature for visual localization that yield impressive results~\cite{milford2012seqslam,cummins2008fab}}. To reduce the storage and increase robustness to environmental changes, one of the research directions is to develop global image descriptors that can be employed to measure image similarity. Usually, a global image descriptor is generated by aggregating hand-crafted features, and the similarity between images are computed by measuring the cosine distances between their descriptor vectors. Apart from hand-crafted features, several works \cite{finetune_hard_samples} utilize convolutional neural networks to extract more discriminative global image features. Recently, the use of binary embedding~\cite{do2018selective} has been demonstrated to be a promising direction, which allows a query image to be quickly compared with other images in the database by measuring hamming distances between binary vectors. \add{However, although binary descriptors enable high compression ratio, each place in the map needs to be associated with one vector, thus the absolute memory must still scale at least linearly with the environment.} \add{Our method allows the above image representations, including binary representations to be further encoded so that sub-linear storage growth can be achieved.}

In the realm of quantization-based approaches, Vector Quantization (VQ)~\cite{gersho2012vector} can be considered as the prime method that lays the foundation for several source coding and retrieval tasks, not only in vision but also in other signal processing applications. 
To improve the effectiveness of VQ, several variants have been proposed~\cite{jegou2011product,ge2013optimized}.  However, in terms of storage scalability, these methods still require the memory to grow linearly with the number of images in the database, since each data instance needs to be encoded by a vector indicating the code-book indexes.
\add{
Several tree-based methods have also been introduced in the literature for scalable visual place recognition such as the vocabulary tree technique proposed in~\cite{nister2006scalable}. However, similar to other quantization approaches, the number of leaf nodes in these approaches must grow linearly with the number of data instances in the database, and each node must store the indexes of the instances belonging to its children.  Therefore, the required memory must grow at least linearly with the amount of training data. Similarly, approaches that are based on traditional decision tree classifiers also require at least linear storage growth, since the number of leaf nodes is proportional to the number of places. In contrast to the above tree-based techniques, our method achieves sub-linear storage growth by encoding training data into a special tree structure. In addition, the training and inference processes of our technique can be performed efficiently, resulting in a storage-efficient algorithm for visual place recognition.
}

%% file: tex/approach.tex
\section{Approach}

This section provides a detailed treatment for the proposed encoding algorithm.
Given a training data set $\cD =\{\bx_i\}_{i=1}^N$ containing $N$ data points, where each data instance $\bx_i$ is a $d$-dimensional image descriptor, the objective of our algorithm is to learn a localization function $f(\cdot)$ such that the location of a query image representation $\bq$ can be best predicted by $f(\bq)$. For simplicity, we assume that the training data has already been sorted and pre-processed as a continuous stream so that the index $i$ of a vector $\bx_i \in \cD$  represents its map location (the terms ``location'' and ``database index'' may also be used interchangeably throughout our discussions).

Traditional approaches for retrieval-based localization typically consider obtaining $f(\bq)$ by nearest neighbor (NN) search (by either direct search or indexing techniques), as the location of a query vector $\bq$ can be approximately identified based on the index of its nearest neighbor in the training dataset. Mathematically speaking, the location $\tilde{i}_{\bq}$ of a query vector $\bq$ can be obtained by
\begin{equation}
    \label{eq:problem_def}
    \tilde{i}_{\bq} = f(\bq) = \arg\min_{i} \|\bq - \bx_i\|_2,
\end{equation}
where each $\bx_i \in \cD$ is a data instance in the training dataset and $\|\cdot\|_2$ denotes the $\ell_2$ Euclidean distance. Although the solution to~\eqref{eq:problem_def} can be obtained by a simple linear scan over the training set, this task becomes computationally expensive when $N$ and the dimensionality $d$ are large, as the complexity for the linear search is $O(Nd)$ (in practice, $d$ is commonly greater than $1024$ and the dataset can contains up to millions of images). As briefly reviewed in Sec.~\ref{sec:introduction} and~\ref{sec:related_work}, this problem is alleviated by the use of approximate nearest neighbor (ANN), in which methods such as ITQ~\cite{gong2013iterative}, PQ~\cite{jegou2011product} and other variants of PQ such as OPQ~\cite{kalantidis2014locally} are used in combination with efficient indexing methods~\cite{invertedindex,invertedmultiindex}. However, due to the fact that the compressed representations of the training data must be stored, the scalability of existing methods remains an issue that needs to be addressed when the size of the training database grows.

In contrast to conventional techniques that compress the original image descriptors to embedding vectors and then conduct the nearest neighbor search~\eqref{eq:problem_def} over the compressed data, we propose a novel mechanism to directly learn the index of a vector in the database. This enables us to jointly learn both the quantization and indexing of the training data. Specifically, from the given training dataset $\cD$, we learn the function $f(\cdot):\bbR^d\mapsto \bbR$ that predicts the location of a feature vector. During inference, the location of a query vector $\bq \in \bbR^d$ can be directly obtained by computing $f(\bq)$. Our method comprises two main steps: Binary Tree Encoding and Tree Learning, which will be discussed in the following sections.

\subsection{Binary Tree Encoding}
\label{sec:binary_tree}
Our tree encoding approach is inspired by the conventional binary search method for one dimensional sorted arrays. To better illustrate the core idea underlying our encoding method, consider an example dataset containing $N=8$ training images as shown in Fig.~\ref{fig:binary_representation} (note that the data instances are indexed from $0$ to $N-1$). For each vector in the training dataset, we denote by $h(\cdot):\bbR^d \mapsto \{0,1\}^b $ the operation that converts its index (in decimal representation) to the corresponding binary representation, where $b$ is the number of bits in the binary number ($N \le 2^b$). As illustrated in Fig.~\ref{fig:binary_representation}, the value of $b$ is $3$, while $h(\bx_0) = 000$, $h(\bx_1) = 001$, \dots, and $h(\bx_7) = 111$. Furthermore, observe that if the values of $h(\bx_i)$ are written vertically as shown in Fig.~\ref{fig:binary_representation}, it is a well-known fact that the arrangement of the $0$ and $1$ bits in the binary representations naturally forms a binary tree, which is further illustrated in Fig.~\ref{fig:tree}. The number of levels in the tree is $b+1$, where the first level (level 0)  is the root node containing all the data points. Except for the leaf nodes, each node in the tree consists of two child nodes: zero-node (or left child, colored in green) and one-node (or right child, colored in yellow in Fig.~\ref{fig:tree}).

Starting from the root node containing the whole training dataset, similar to conventional binary search, the vectors are recursively partitioned as follows. At a particular level $l$, the left child (zero-node) contains all data points from its parent node having the $l$-th bit in their binary representations being $0$ (note that the indexes of the bits start from $1$). The same applies to the right child (one-node), but for data points from the parent node having the $l$-th bit in their binary representations being $1$. Take for instance level 1 shown in Fig.~\ref{fig:tree}, the left child node contains $\{\bx_0, \dots,\bx_3\}$ because the first bit of these data points are $0$, while $\{\bx_4, \dots,\bx_7\}$ belongs to the right child since their first bit are $1$. The same division procedure applies to other nodes further down the tree.

\subsection{Tree Training}
\label{sec:tree_training}
Next, based on the introduced tree structure, we describe two training schemes to learn the parameters for the localization task. The pros and cons of each training scheme will also be discussed in detail and validated in the experiments.

\subsubsection{Full Training} This scheme follows closely the traditional binary search, in which each parent node in the tree is associated with a binary classifier. Formally, at node $j$ (the nodes are numbered from top to bottom, left to right, starting from $0$, as illustrated by red numbers in Fig.~\ref{fig:tree}), a classifier $\tilde{f}_j(\cdot)$ is learned, where the training data of $\tilde{f}_j(\cdot)$ consist of data points belonging to that particular node. The training label of each data point $\bx_i$ is the $(l+1)$-th bit of $h(\bx_i)$, where $l$ is the level in the tree to which node $j$ belongs. In Fig.~\ref{fig:tree}, for example, the training data for node $2$ includes $\{\bx_4,\cdots,\bx_7\}$, and the label vector -- since node $2$ belongs to level $1$ -- is the values of the second bit in the indexes' binary representations, i.e., $[0,0,1,1]$, as shown in Fig.~\ref{fig:binary_representation}. 


Once all the node classifiers are successfully trained, given a query vector $\bq$, its location is identified by a binary-search-like inference process. Specifically, we pass $\bq$ down the tree, starting from the root node (node $0$), until reaching a leaf node. At termination, the index of the data point associated with the final leaf node is returned as the location of $\bq$. Intuitively, at node $n$, we use the trained classifier $\tilde{f}_n(\cdot)$ to predict whether $\bq$ should be sent to the left or right child. This evaluation process is recursively repeated until a leaf node is reached. 

\subsubsection{Compressed Training} 
Using the full training scheme, the number of classifiers that need to be trained is $O(2^{b})$, where $b$ is the number of bits in the binary representations. Therefore, the memory complexity is $O(N)$, since $N\le 2^b$ (assuming that the memory complexity of each classifier is $O(1)$). For example, in Fig.~\ref{fig:tree_encoding}, with $N=8$, seven different classifiers associated with node $0$ to node $6$ must be trained. Thus, although the time complexity of the inference is $O(\log N)$, this full training scheme requires the storage to scale at least linearly with $N$. In this section, we show an improved scheme that requires sub-linear storage complexity.



This compressed training scheme aims to reduce the overall required memory such that the memory is sub-linear with respect to the database size $N$. Particularly, rather than training $O(2^b)$ classifiers as in the full training scheme, we propose to train only $b$ binary classifiers corresponding to $b$ levels of the tree (from level $1$ to level $b$). This scheme is based on the observation in Fig.~\ref{fig:binary_representation} that the prediction of each bit in the binary representation can be cast as a binary classification problem. \add{The additional advantage of this training scheme is that $b$ classifiers corresponding to $b$ levels of the tree can be trained in parallel}. Specifically, with a dataset requiring $b$ bits in the binary representations, we train $b$ binary classifiers $\{g_j(\cdot)\}_{j=1}^{b}$, where $g_j$ is associated with level $j$ of the tree. The training data of each classifier $g_j(\cdot)$ is the whole training dataset $\{\bx_i\}_{i=1}^N$, while the binary label for each data instance $\bx_i$ is the $j$-th bit of $h(\bx_i)$. During inference, the combination of results obtained from $g_j(\bq)$ gives the predicted location of a query vector $\bq$. In particular, the predicted location $\tilde{i}_\bq$ is computed by
\begin{equation}
    \tilde{i}_\bq = h^{-1}([g_1(\bq) \dots g_{b}(\bq) ]),
\end{equation}
where $h^{-1}(\cdot)$ denotes the conversion of a binary number into its corresponding decimal value.

Using this compressed training scheme, the memory complexity is $O(\log N)$, since only $b$ classifiers are required. Several machine learning approaches can be employed to train the binary classifiers as previously discussed. In this work, to maintain $O(\log N)$ memory complexity, the classifier type is chosen such that its required memory is independent of $N$. Therefore, we employ linear Support Vector Machines (SVM) as binary classifiers, as each binary SVM requires to store a $(d+1)$ dimensional hyper-plane. 
\add{Since we employ SVM for tree encoding, our training schemes aim to minimize the number of places that are misclassified instead of the information gain metric that is commonly employed by traditional decision tree classifiers.}

\vspace{-0.1cm}
\subsection{\add{Dimension Reduction}}
\label{sec:subset_selection}
Observe that the introduced training schemes use the original feature vector $\bx_i \in \bbR^d$ to train the classifiers. Consequently, the overall memory depends on the dimensionality of the image descriptors. In this section, we introduce a feature selection technique to further reduce the storage footprint by using only a small subset of the feature map for training. This technique can be applied before training any binary classifier discussed in Sec.~\ref{sec:tree_training}.

\begin{figure}[ht]
    \centering
    \includegraphics[width = 0.9\columnwidth]{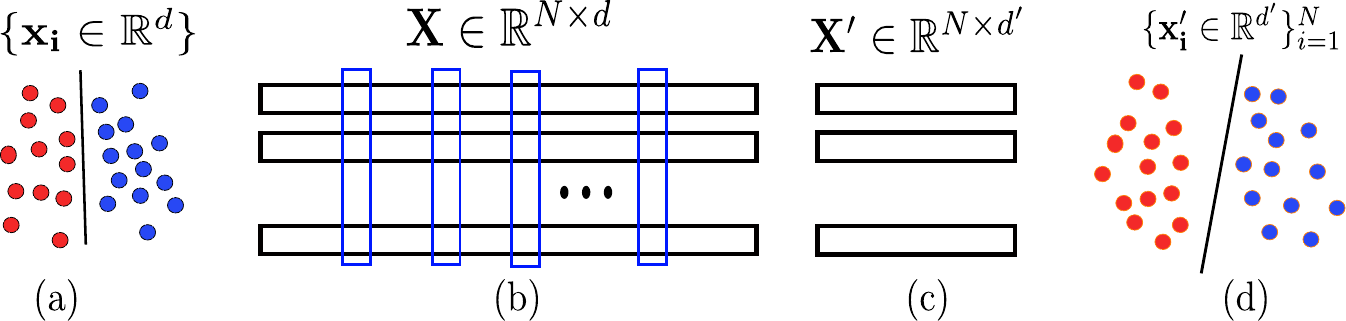}
    \caption{Illustration of feature map selection. (a) Original database with two labels that may not be properly clustered. (b) A number of columns in the original database $\bX$ are selected to form a new matrix $\bX'$ (c). (d): The new dataset form by dimension reduction that are separated. }
    \label{fig:column_selection}
\end{figure}

The core idea is that, rather than using the original feature vectors $\{\bx_i \in \bbR_d\}_{i=1}^N$ as input training data, we extract in each vector $\bx_i$ a subset containing $d'$ ($d' \le d$) elements to form new training data points $\{\bx_i' \in \bbR^{d'}\}_{i=1}^N$. Let $\bX \in \bbR^{N\times d}$ be the matrix containing all the training data where each row of $\bX$ is a data instance, and let $y_i\in\{0,1\}$ represent the label of $\bx_i$. \add{Dimension reduction involves selecting $d'$ columns of $\bX$ ($d' \le  d$) such that the linear separability of the data points in the reduced subspace (i.e., $\bbR^{d'}$) 
is maximized, which is illustrated in Fig.~\ref{fig:column_selection}. }
\add{Note that our dimension reduction scheme is different from traditional unsupervised dimensional reduction strategies such as Principal Component Analysis (PCA) since PCA only projects data into principal components such that the variances are maximized, while the labels are not taken into account. Therefore, the use of PCA does not maximize the linear separability as proposed by our technique.}

Our subspace extraction technique is inspired by several works on feature selection for K-Means clustering~\cite{witten2010framework}. However, here we consider a simplified version in which we assume that the dataset has been divided into two clusters $\bS_0$ and $\bS_1$, where $\bS_l = \{\bx_i \in \cD | y_i = l\}$, $l\in \{0,1\}$, \add{where $y_i$ refers to the label associated with the data point $\bx_i$ for a particular tree level} and we only solve the feature selection problem. 

Formally, the task of selecting $d'$ columns can be done by learning a weight vector $\bw \in \bbR^d$ that solves the following problem~\cite{witten2010framework}
\begin{equation}
    \label{eq:w_learning}
    \begin{aligned}
    & \min_{\bw \in \bbR^d} &&\sum_{l = 0}^{1} \sum_{j = 1}^{d} \sum_{\bp,\bq \in \bS_l, \bp \neq \bq} \bw_j (\bp_j - \bq_j)^2,\\
    & \text{s.t.} && \bw^2_1 + \dots + \bw^2_d \le 1, 
    |\bw_1| + \dots + |\bw_d| \le s, 
     \bw_j \ge 0 \;\; \forall j,
    \end{aligned}
\end{equation}
where $s$ is a tuning parameter. In our experiments, we choose $s=0.1$. Intuitively, by optimizing the objective of~\eqref{eq:w_learning}, we are selecting $d'$ columns of $\bX$ so that the within-cluster sum of squares (WCSS) of the data points are minimized. The weight $\bw_i$ of each column reflects the importance of that particular column in the selection process.
Following~\cite{witten2010framework}, we employ soft-thresholding  technique to solve~\eqref{eq:w_learning}. Once the vector $\bw$ is obtained, $d'$ elements in $\bw$ with the largest weights are selected as the indexes of extracted columns in the database matrix $\bX$.

\subsection{Training Data Clustering with Sequence Summarization}
\begin{figure}[ht]
    \centering
    \includegraphics[width=0.8\columnwidth]{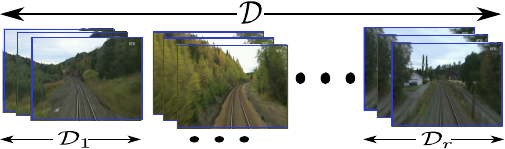}
    \caption{Illustration of sequence summarization. The original training data $\cD$ is partitioned into $r$ regions, which form $r$ new training datasets $\{\cD_k\}_{k=1}^r$.}
    \label{fig:video_summ}
\end{figure}

In several localization applications, a vehicle may drive through several regions, where images captured in each region are relatively similar~\cite{mcmanus2015learning}. Thus, to improve the localization, we propose to divide the training dataset $\cD$ into $r$ sub-datasets $\{\cD_k\}_{k=1}^r$, where each dataset $\cD_k$ contains training images in region $k$, which is illustrated in Fig.~\ref{fig:video_summ}. In our work, this region division is achieved by a novel adaptation of the video summarization technique proposed in~\cite{potapov2014category}. After obtaining $r$ different datasets, each dataset is individually encoded using the method described in Sec.~\ref{sec:tree_training}. Particularly, for each sub-dataset $\cD_i$,  we learn a function $f_i(\cdot)$ that serves as a sub-localizer for that particular region.

As the original dataset is now partitioned, during the inference process, the algorithm needs to know which region $\cD_k$ the vehicle is currently in so that the right classifier $f_k(\cdot)$ is used. In real-world scenarios, one can make use of information from GPS sensors to achieve this coarse estimation. In our work, for completeness, we propose to learn the region information by an additional multi-class SVM classifier $\mathcal{R}(\cdot)$. The training data of $\mathcal{R}(\cdot)$ is the original dataset $\cD$ and the label of each instance is the region in which it belongs. \add{In general, the use of data clustering can be considered as a hybrid of the Full Training and Compressed Training schemes introduced in Section~\ref{sec:tree_training}, where Full Training is used for the first level, and Compressed Training is used within each Full Training leaf.}

%% file: tex/results.tex
\section{Experimental Results}

\begin{figure*}[h]
    \centering
    \begin{subfigure}{\textwidth}
    \centering
    \includegraphics[width=0.28\textwidth]{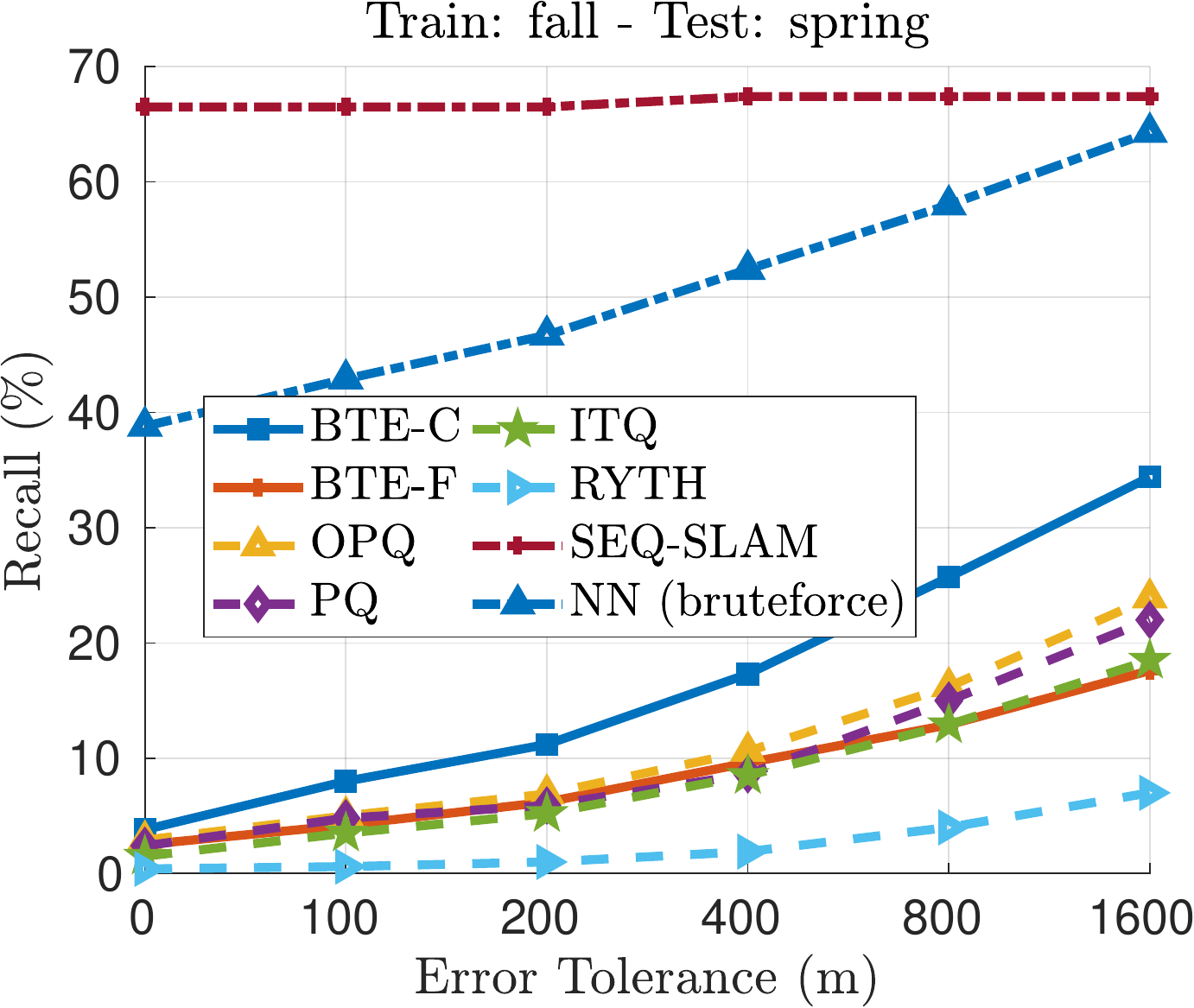}
    \includegraphics[width=0.28\textwidth]{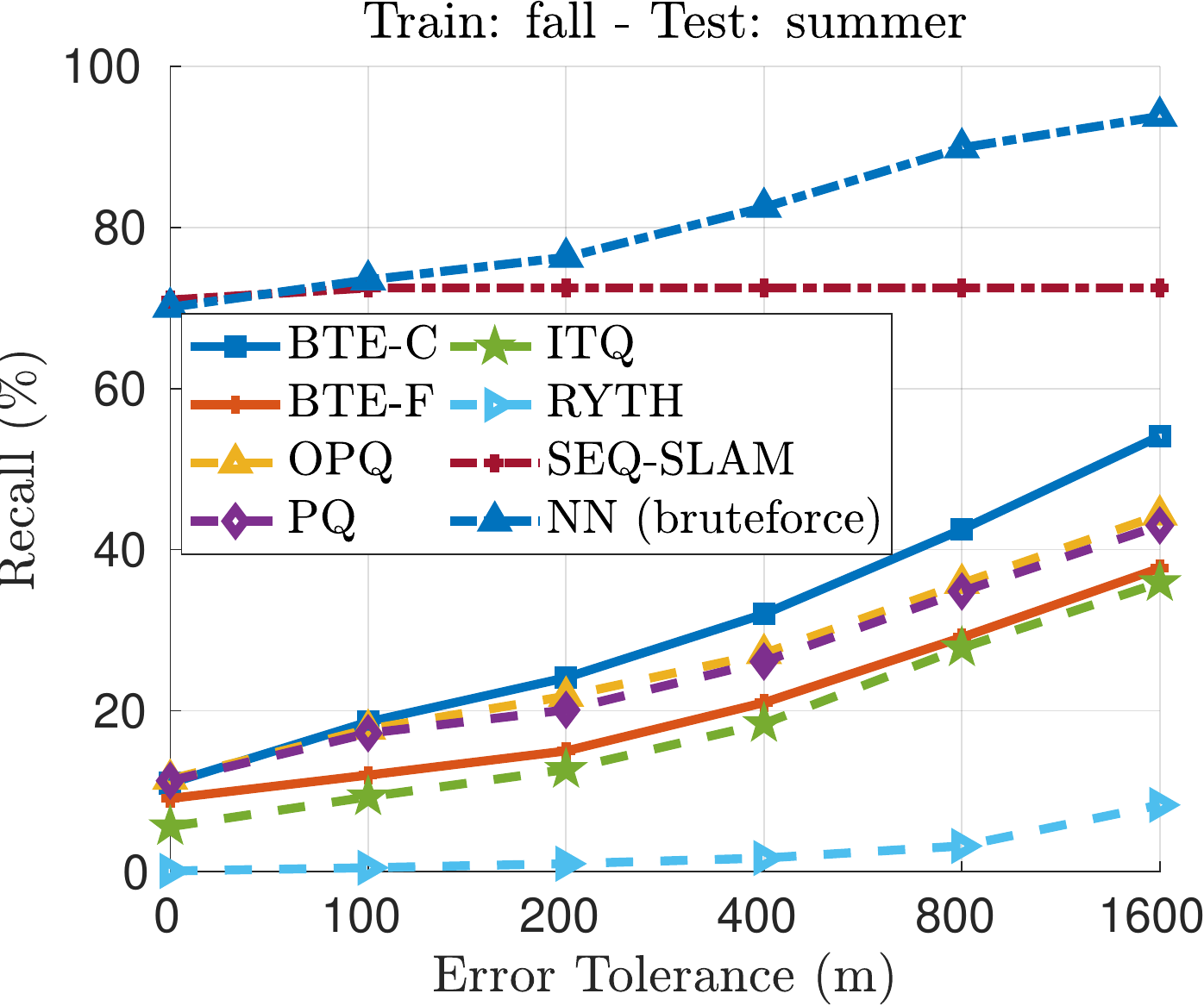} 
    \includegraphics[width=0.28\textwidth]{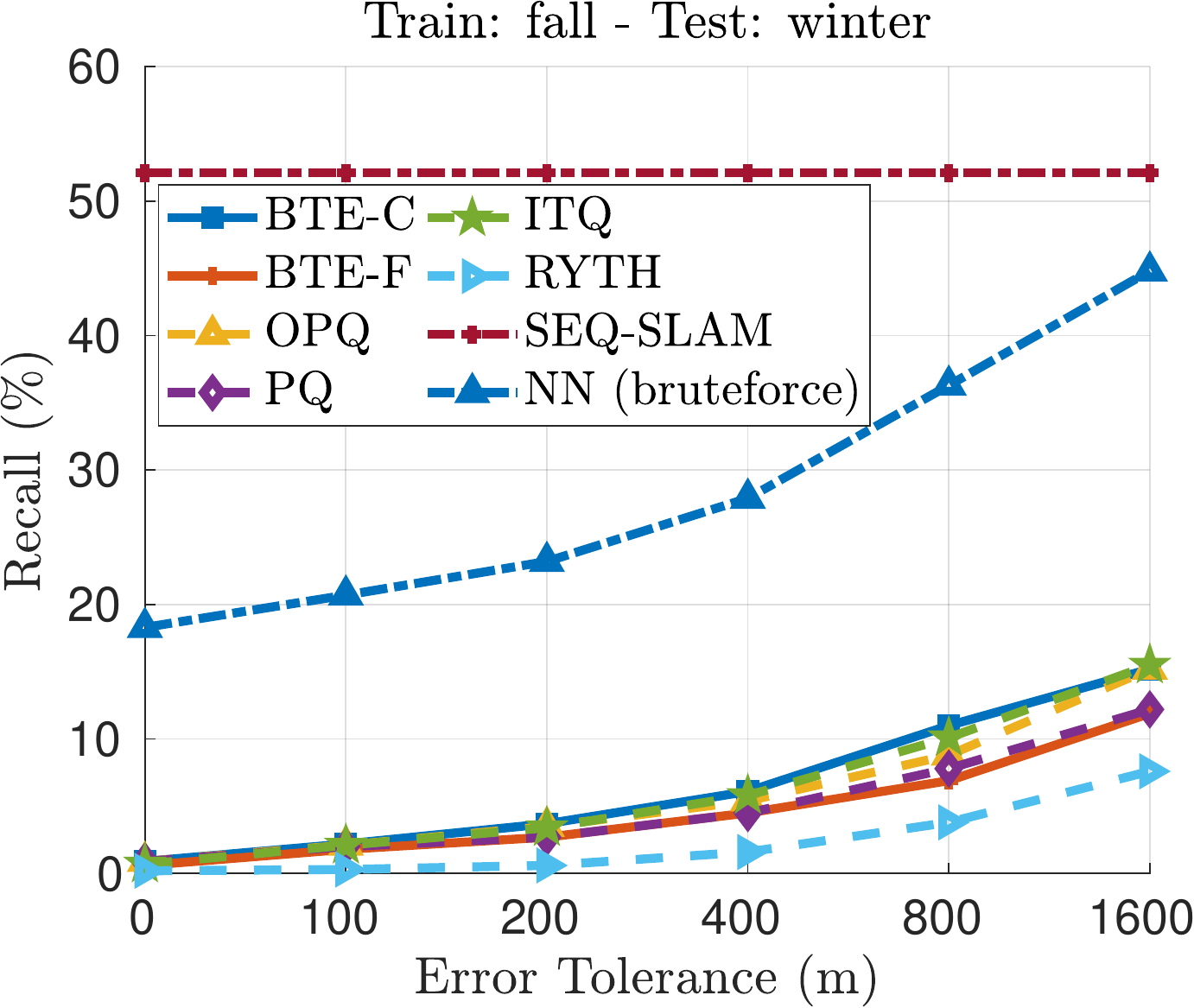}
    \caption{Performance on Nordland Dataset trained on Fall and tested on Spring, Summer and Winter.}
    \label{fig:accuracy_fall}
    \end{subfigure}
    \begin{subfigure}{\textwidth}
    \centering
    \includegraphics[width=0.28\textwidth]{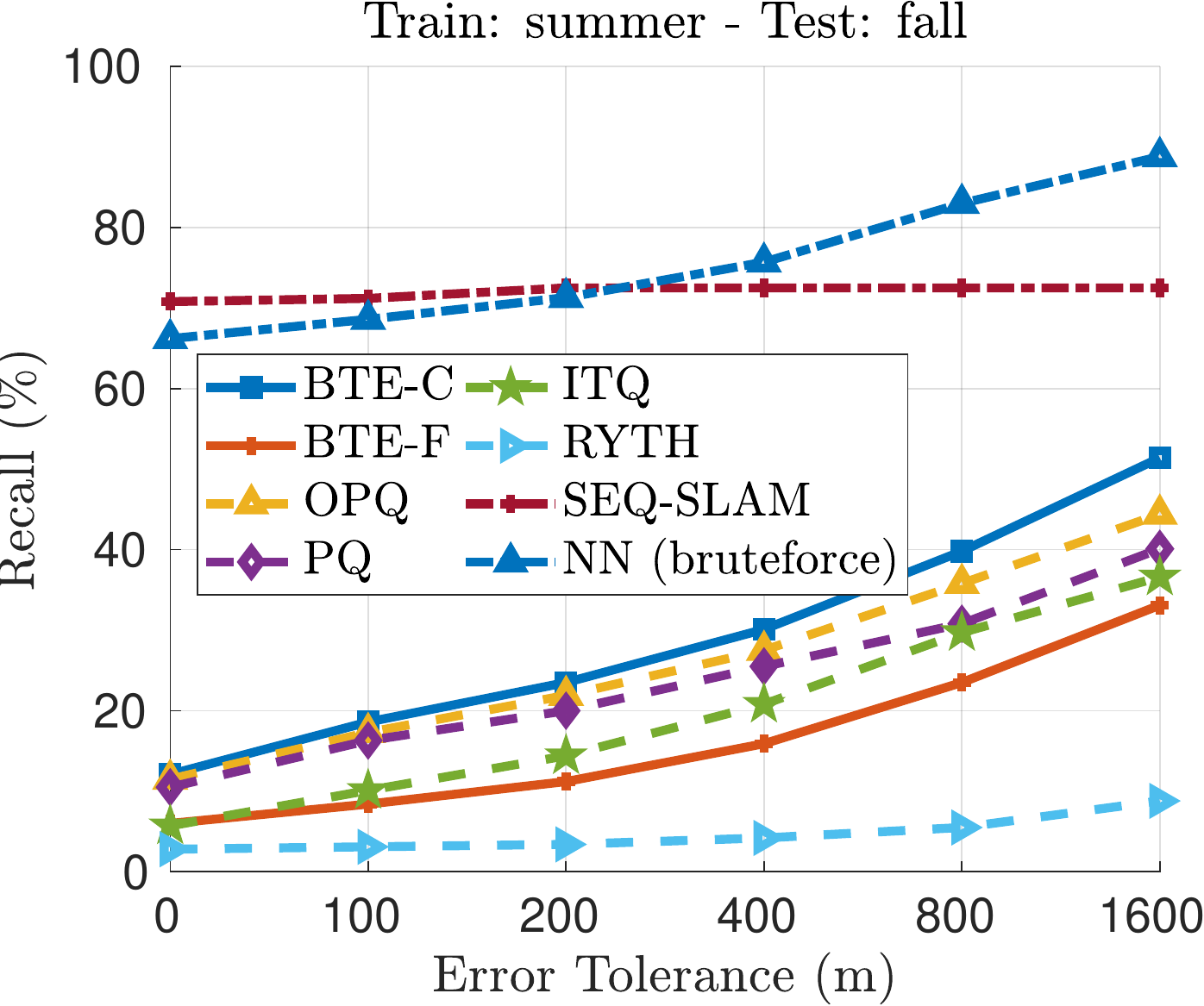}
    \includegraphics[width=0.28\textwidth]{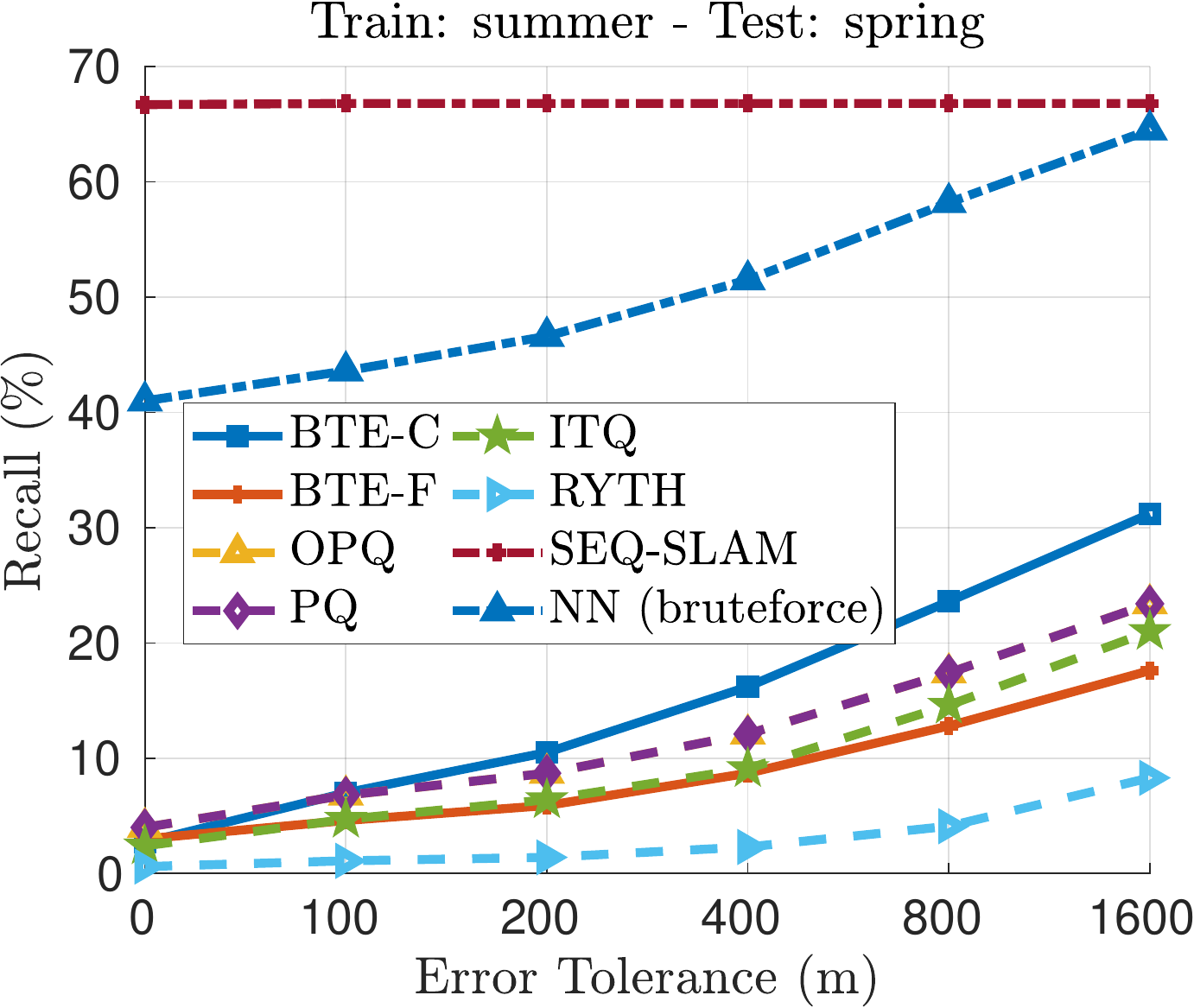} 
    \includegraphics[width=0.28\textwidth]{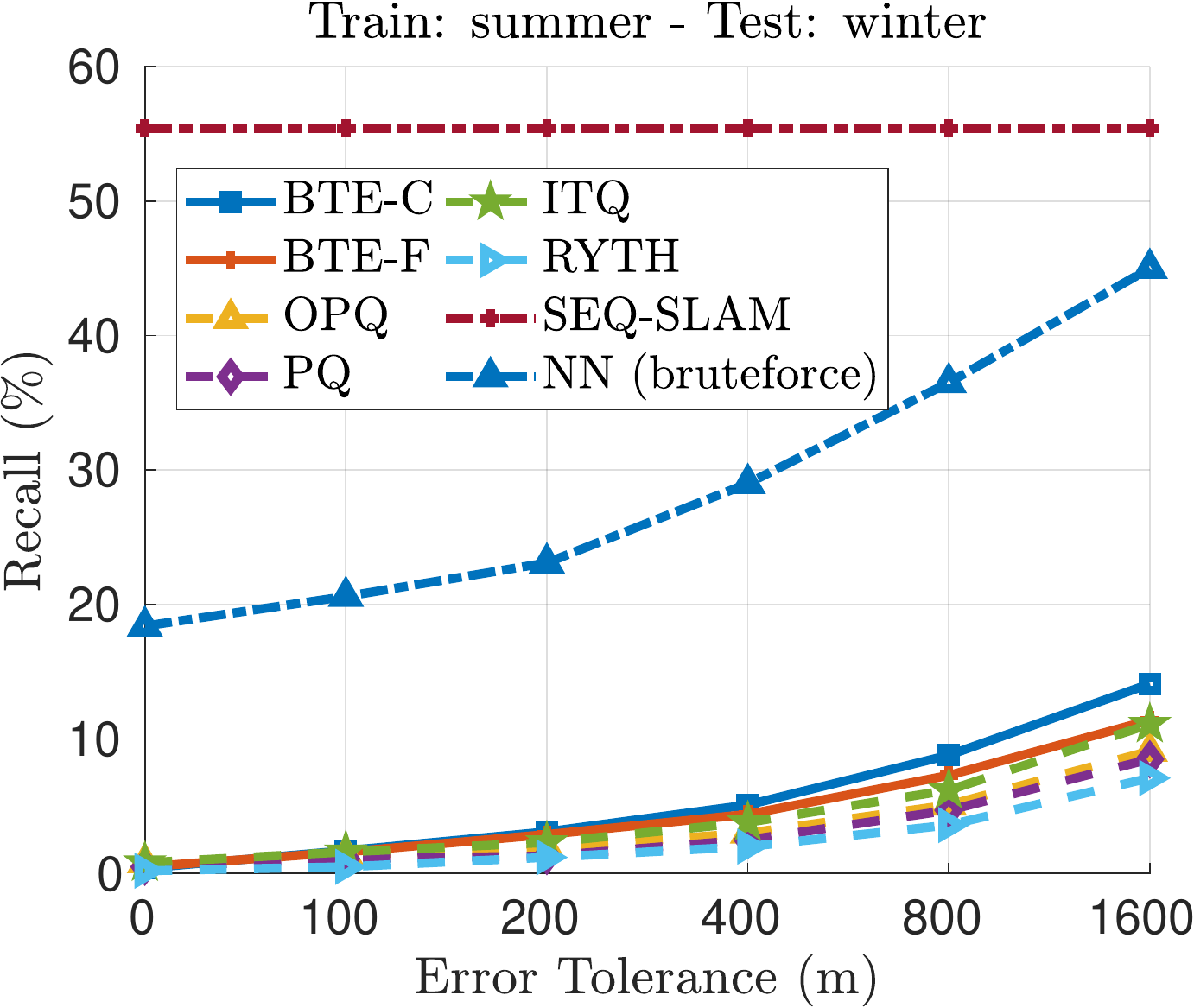}
    \caption{Performance on Nordland Dataset trained on Summer and tested on Fall, Spring and Winter.}
    \label{fig:accuracy_summer}
    \end{subfigure}
    \caption{Localization accuracy of benchmarking methods on Nordland Dataset (with $N=2000$) with $2$MB storage.}
    \label{fig:accuracy}
\end{figure*}

In this section, we conduct several experiments to evaluate the performance of our proposed encoding methods compared to existing state-of-the-art approaches. In addition, we provide a detailed analysis on the algorithm's characteristics under the effect of varying parameter settings, e.g., the reduced dimension $d'$, the number of regions $r$, and the scalability of the method for increasing database sizes. A Python implementation of our algorithm is provided at: https://tinyurl.com/y5y9x8nj
\subsection{Dataset and Evaluation Metric}
We benchmark our algorithm on the Nordland dataset\footnote{https://nrkbeta.no/2013/01/15/nordlandsbanen-minute-by-minute-season-by-season/}, which is captured from a front-facing camera installed at the front of a train running for around 10 hours long. Four video sequences are collected through four seasons of the year: fall, summer, spring and winter (for brevity, we use the name of the season to refer to the corresponding sequence). 
For each sequence, we extract a subset containing 8,200 frames covering the whole traversal (with identical distance between frames) for our experiments. Each frame in the sequence is described by DenseVLAD~\cite{torii201524}, which is among the state-of-the-art descriptors for visual place recognition, resulting in feature descriptors having dimensionality of $d=4096$. 

Following~\cite{torii201524}, we evaluate the performance of all the methods by measuring the percentage of correctly localized images  (recall rates), \add{with a tolerance of $t$ meters}. Specifically, a query vector $\bq$ is considered to be correctly localized if the predicted location is $t$ meters away from the ground-truth location. \add{For the Nordland dataset, an error of 200 meters correspond to approximately 10 frames in our database}. 

Since we focus on demonstrating the efficiency in terms of storage of the methods, we compare our  methods (including Full Training and Compressed Training scheme) against several state-of-the-art quantization techniques that are commonly used in retrieval-based localization, including ITQ~\cite{gong2013iterative}, PQ~\cite{jegou2011product} and OPQ~\cite{ge2013optimized}, \add{where a query image is localized using approximate nearest neighbor search}. Note that to obtain the best performance of these quantization methods, we use exact nearest neighbor search, i.e., linear scan over the code vectors. In addition, we also compare our algorithm with the Rhythmic Representation method (RYTH) proposed by Yu et al.~\cite{yu2018rhythmic}, which is the first method to achieve sub-linear storage growth. To obtain baseline results produced by non-compress methods, we also compare our method against SeqSLAM~\cite{milford2012seqslam} and the localization results obtained from brute-force nearest neighbor search of the descriptor vectors.   

\subsection{Localization Accuracy}
\begin{figure}
    \centering
    \includegraphics[width=0.85\columnwidth]{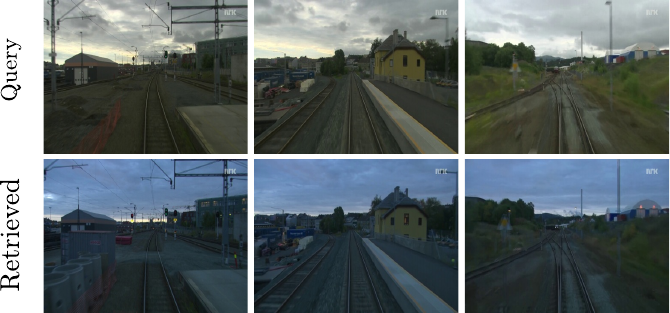}
    \caption{Example of correctly localized queries. Top: Query images from Summer. Bottom: Images in the Fall sequence that are correctly retrieved. Note that images are for visualization purpose, as our system does not store images.}
    \vspace{-0.7cm}
    \label{fig:my_label}
    
\end{figure}

This section evaluates the localization accuracy of the methods with a fixed amount of memory, in which we will show that our system achieves competitive localization accuracy compared to state-of-the-art quantization approaches and significantly outperforms the sub-linear storage algorithm proposed in~\cite{yu2018rhythmic}.

In this experiment, for each sequence, we use the first 2000 frames in the set of extracted frames \add{(we test with 2000 frames so that RYTH~\cite{yu2018rhythmic} achieves meaningful results, since its recall is almost zero when $N>2000$. Experiments with larger $N$ will be conducted in the following sections)}. We train all methods on the Fall sequence and test the localization on the three remaining sequences (i.e., Spring, Summer and Winter). To simulate the settings with limited storage budget, we adjust the parameters such that the required storage footprint of each method is not greater than 2MB. Fig.~\ref{fig:accuracy_fall} shows the results, where we plot the accuracy of the methods with the error tolerance $t$ ranges from $0$ to $1600$ meters. Note that for our method (BTE), we also show the results for both the full training scheme (BTE-F) and the compressed training scheme (BTE-C).  Fig.~\ref{fig:accuracy_summer} shows the result for the same experiment, but the methods are trained on Summer and tested on the three remaining sequences.

Under a tight memory constraint ($\le 2$MB), our compressed training scheme (BTE-C) significantly outperforms other competitors. Moreover, as shown in Fig.~\ref{fig:accuracy}, although achieving sub-linear storage growth, RYTH~\cite{yu2018rhythmic} does not provide satisfactory localization results for datasets with visually changing conditions. In contrast, our approach (BTE-C) has demonstrated a significant improvement in terms of accuracy, while also achieves a sub-linear storage growth. This shows the effectiveness of our proposed tree encoding in combination with the feature selection and sequence summarization scheme. 
\add{Compared to the baseline results (without compression) obtained from brute-force nearest neighbor search and SeqSLAM, our method achieves a compression ratio of $6\times 10^{-5}$ and experiences drops in the overall performances. However, the achieved performance shown in Fig.~\ref{fig:accuracy} are still useful in several applications.}

From the results plotted in Fig.~\ref{fig:accuracy}, we also show that the full training scheme (BTE-F) is not necessary, as the results of BTE-C is better than that of BTE-F by a large margin. This is because in order to satisfy the memory constraint, the dimensionality of the training data in each node must be reduced to an extend that it no longer captures the data association. This also reveals the disadvantages of methods with linear storage growth to deal with massive amount of training data under a limited memory budget. Similarly, for quantization methods such as OPQ, PQ and ITQ, besides the fact that each training data instance must be associated with a code vector (which scales linearly with the training size), the code-books, each containing many code-words must also be stored. It has been empirically shown that in order to achieve better accuracy, one should use the large code-books. However, if the memory is limited, the number of code-books and code-words must be reduced, resulting in the degradation in overall performance.

As BTE-C has proven to be much better than BTE-F, in the sequel, we focus on the evaluation of BTE-C and refer to this method as BTE for brevity.
\subsection{System Scalability}
\begin{figure}[ht]
    \centering
    \includegraphics[width = 0.45\columnwidth]{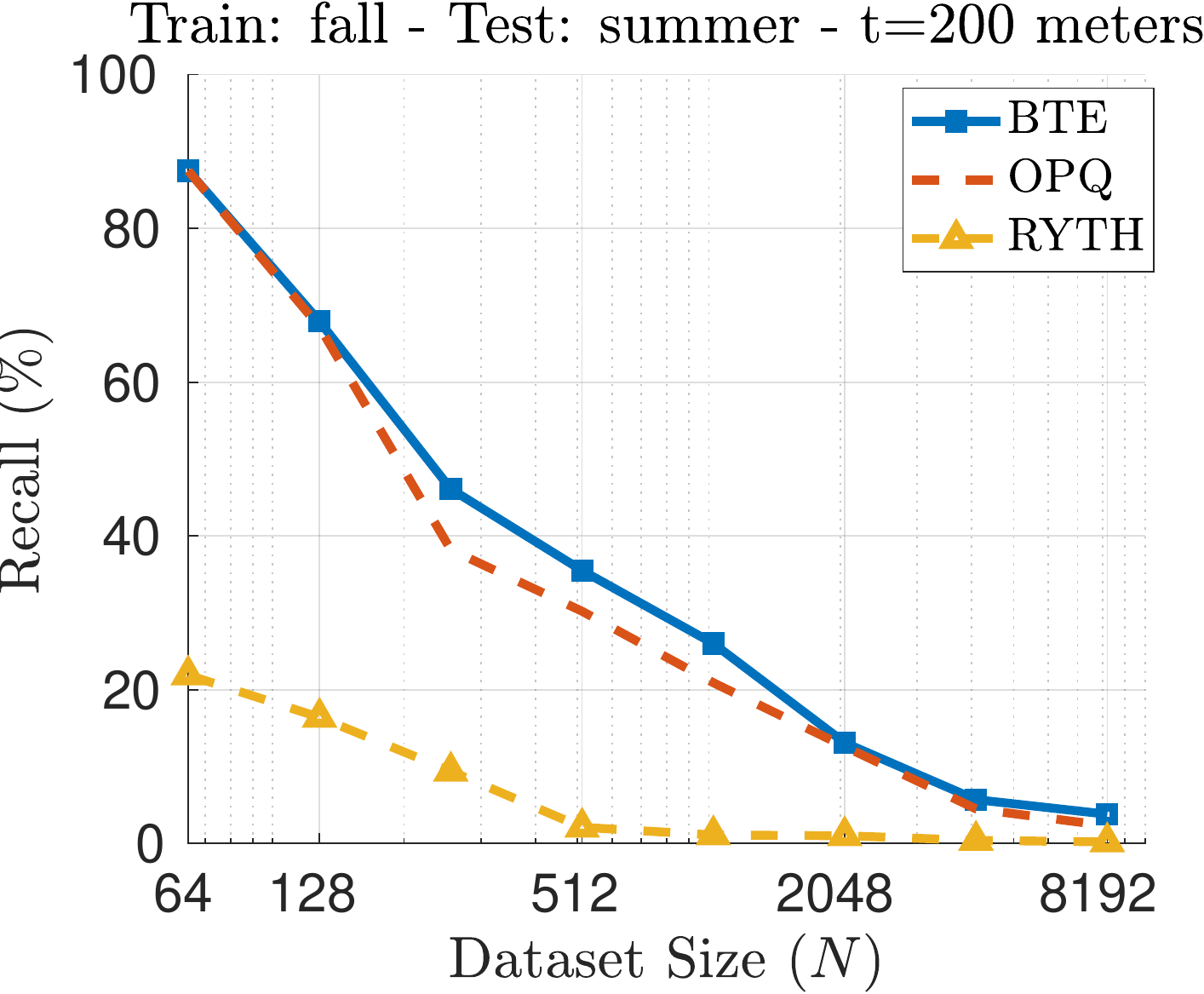}
    \includegraphics[width = 0.45\columnwidth]{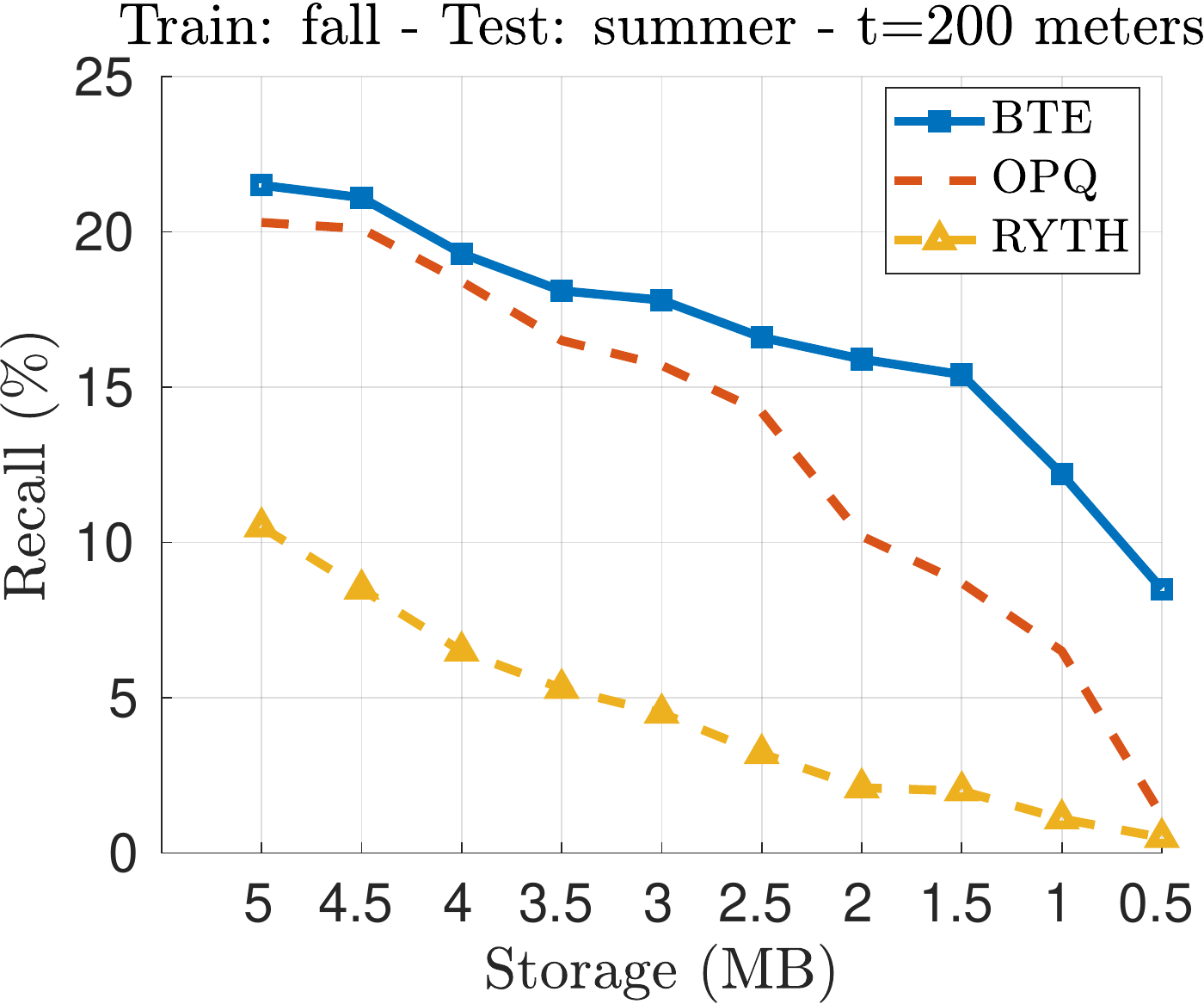}
    \caption{Analysis of the scalability of our method (BTE), OPQ and RYTH~\cite{yu2018rhythmic}, with a error tolerance of $t=200$ meters. Left: Recall with varying database size $N$ when the storage is fixed to $0.5$MB. Right: Recall with varying storage budget when  $N = 8200$.}
    \label{fig:scalability}
\end{figure}
In this analysis, we show how well the methods perform as the number of training data grows with a fixed amount of storage. We enforce a tight memory constraint of $\le$0.5MB, then execute the methods with increasing training and testing sizes when the number of training and testing images $N$ increases from  $64$ to $8200$ frames. The results are plotted in Fig.~\ref{fig:scalability} (left), where we compare BTE with OPQ~\cite{ge2013optimized} and RYTH~\cite{yu2018rhythmic}. As can be seen, when $N$ is small, our method provides equivalent results compared to OPQ. However, as $N$ increases, OPQ's performance starts to drop, and our method is more accurate. 
\add{
We repeat the experiment with a fixed database size of $N=8200$ and vary the storage budget from $5$MB to $0.5$MB. The results are shown in Fig.~\ref{fig:scalability} (right), where our method demonstrates better scalability compared to OPQ and RYTH.
}

Fig.~\ref{fig:scalability} also shows that BTE achieves not only better localization accuracy but also better scalability than RYTH. In fact, the accuracy of RYTH drops much faster and when $N>2000$, its accuracy is almost zero. This also confirms that our method serves as a better alternative for RYTH in applications that require sub-linear storage growth.

\subsection{Sequence Filtering}
In several visual SLAM applications, it has been shown that the use of Sequence Filtering (SF) can significantly improve the localization accuracy by utilizing the temporal information between consecutive frames~\cite{milford2012seqslam}. Our proposed encoding method is also amenable to such filtering approach. 
In particular, our results can be corrected by utilizing the predicted locations for consecutive frames within a running window with the size of $a$ frames. When a new query frame $\bq$ arrives at the system, its predicted location $i_{\bq}$ is compared against the median predicted location $m_{\bq}$ of $2a$ previous frames. If the predicted location $i_\bq$ is more than $(a+1)$ frames away from the median $m_{\bq}$, the prediction is considered wrong and corrected by $i_\bq = m_\bq + a + 1$. 
\begin{figure}[ht]
    \centering
    \includegraphics[width=0.49\columnwidth]{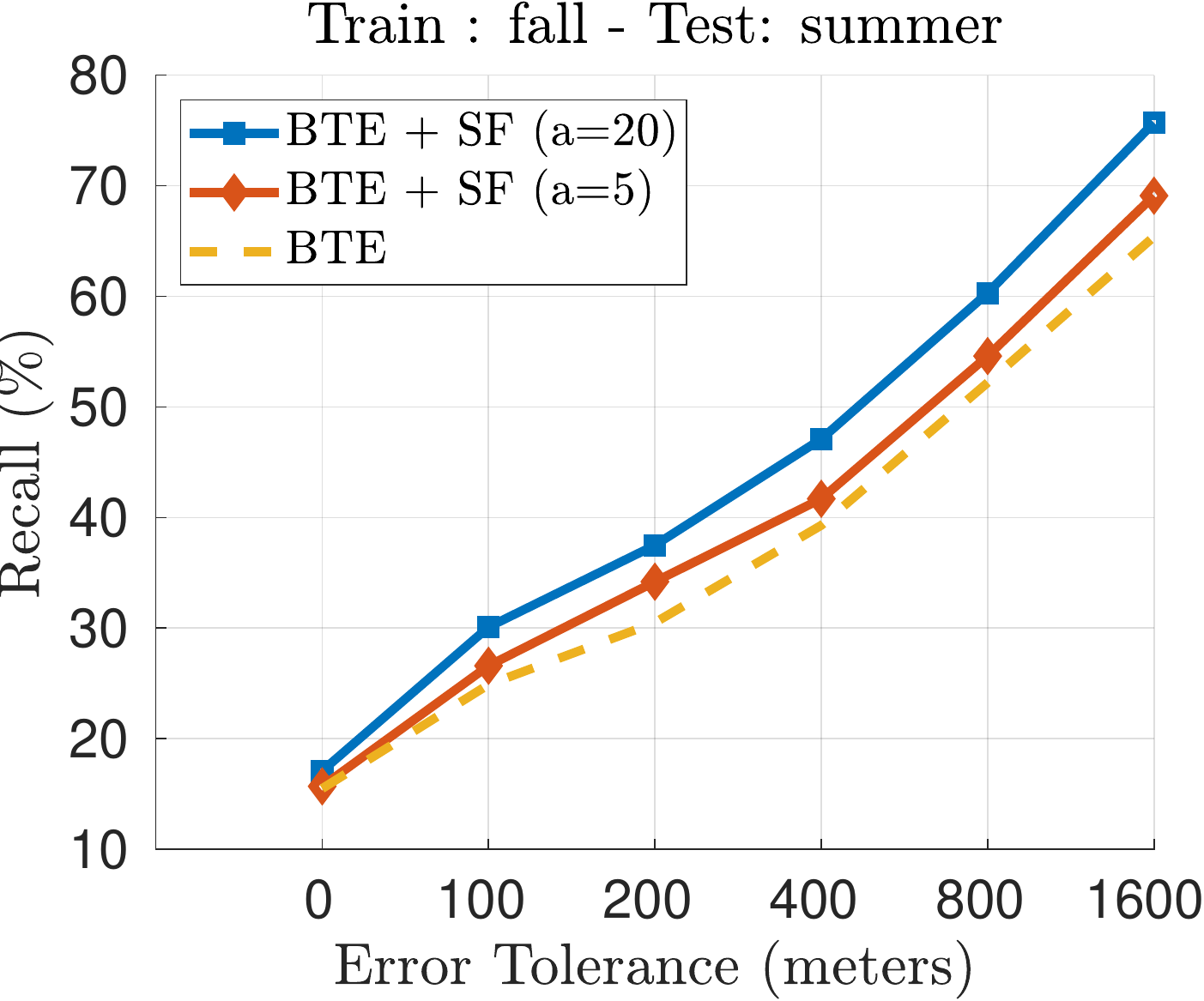}
    \includegraphics[width=0.49\columnwidth]{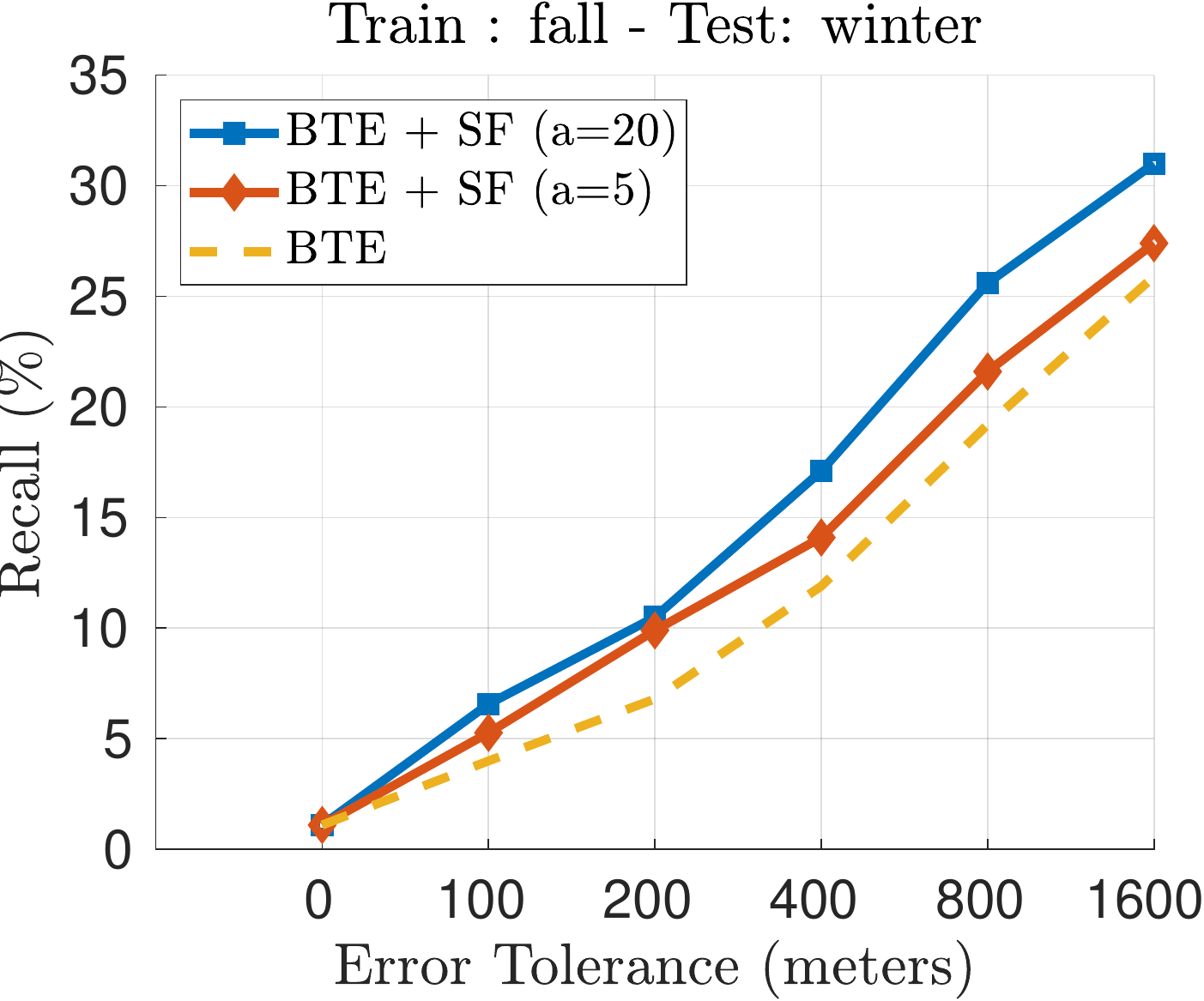}
    \caption{Performance of BTE with sequence filtering for a dataset containing $5000$ frames with storage memory of $1MB$.}
    \label{fig:sequence_filter}
    \vspace{-0.2cm}
\end{figure}
Fig.~\ref{fig:sequence_filter} shows the results of BTE with and without filtering for $N=5000$ frames and two different window sizes of $a=5$ and $a=20$ frames. Observe that the use of temporal information obtained from sequence filtering has considerably improved the localization results. Note that the use of a sequence filter does not increase the total required storage, as only temporary memory to store $2a$ frames is needed during inference. SF provides an additional user-determinable trade off, as one may reduce the total required storage and accept a poorer single frame accuracy, but it can be compensated during run-time by SF with large window size.

\subsection{Parameter Analysis}
\begin{figure}[ht]
    \centering
    \includegraphics[width = 0.49\columnwidth]{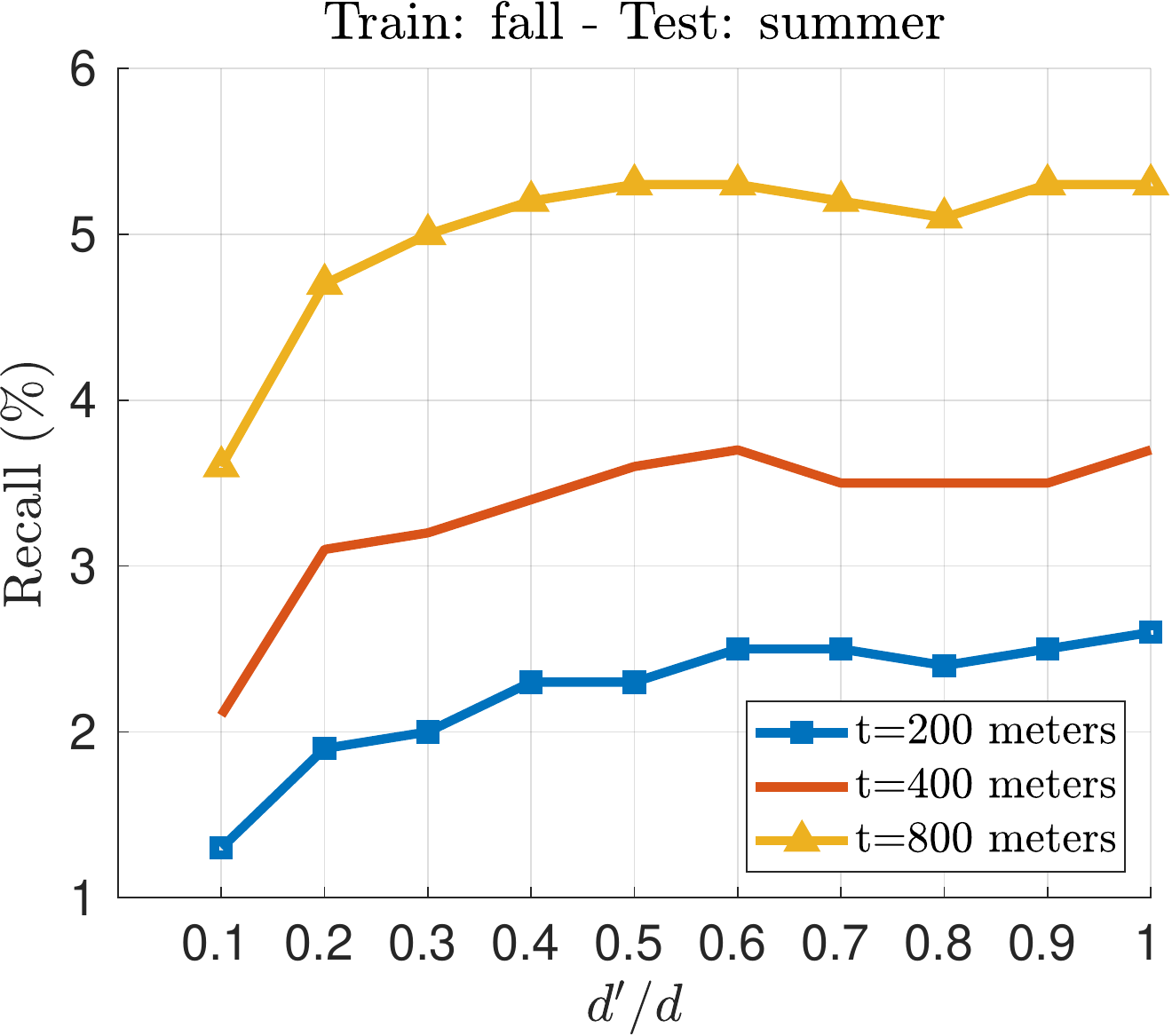}
    \includegraphics[width = 0.49\columnwidth]{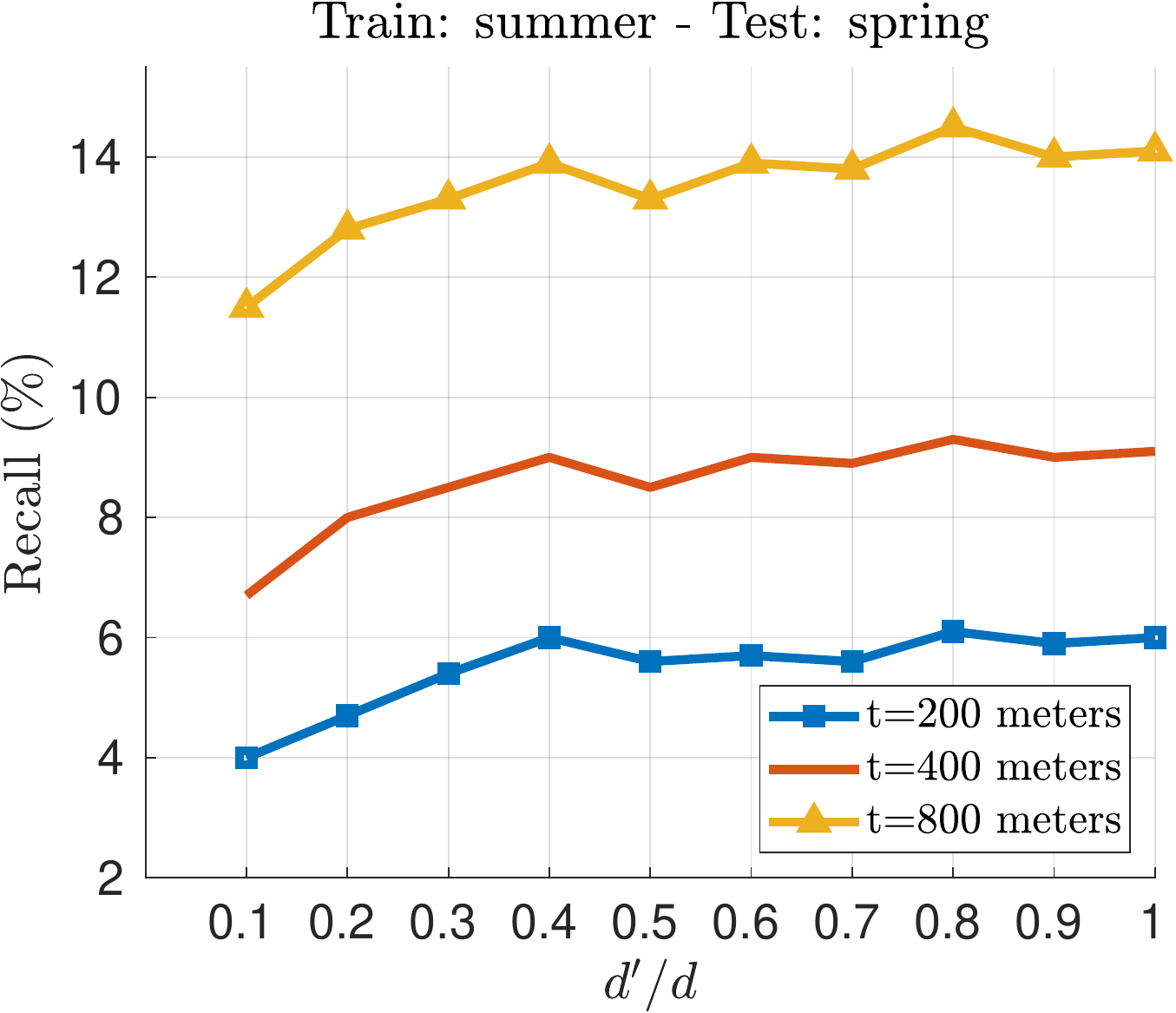}
    \caption{Performance of the compressed training scheme with varying values of $d'$ when $r=1$.}
    \label{fig:dim_analyze}
\end{figure}

Next, we analyze the performance of our algorithm under varying parameter settings. The experiments in this section are  conducted on the extracted $8200$ frames in each sequence. In the first experiment, we evaluate the performance of the system under the effect of feature selection, where the ratio of the reduced dimension over the original dimension $c=\frac{d'}{d}$ varies from $0.1$ to $1$. The results for different error tolerance are plotted in Fig.~\ref{fig:dim_analyze}. Observe that our method achieves good localization results even when $d' = 0.1d$, due to the ability our feature selection scheme to obtain good features to train the classifiers, without the need to use the whole feature vector as proposed in~\cite{yu2018rhythmic}. This is among the key factors allowing our system to achieve much better performance while the memory footprint is significantly smaller than~\cite{yu2018rhythmic}.

As shown in Fig.~\ref{fig:dim_analyze}, when $d'$ increases, the system achieves better localization results. However, observe that starting from the point where $d'=0.4d$, the increase in performance is no longer significant. This demonstrates that our scheme is able to exploit redundancy in the feature vectors, and also further explains why our proposed method is able to achieve both high compression ratio and high localization accuracy. 
\add{The experiment with the number of clusters $r$ fixed to $1$ also demonstrates the applicability of our tree encoding mechanism even without clustering of the training data.}

\begin{figure}
    \centering
    \includegraphics[width = 0.49\columnwidth]{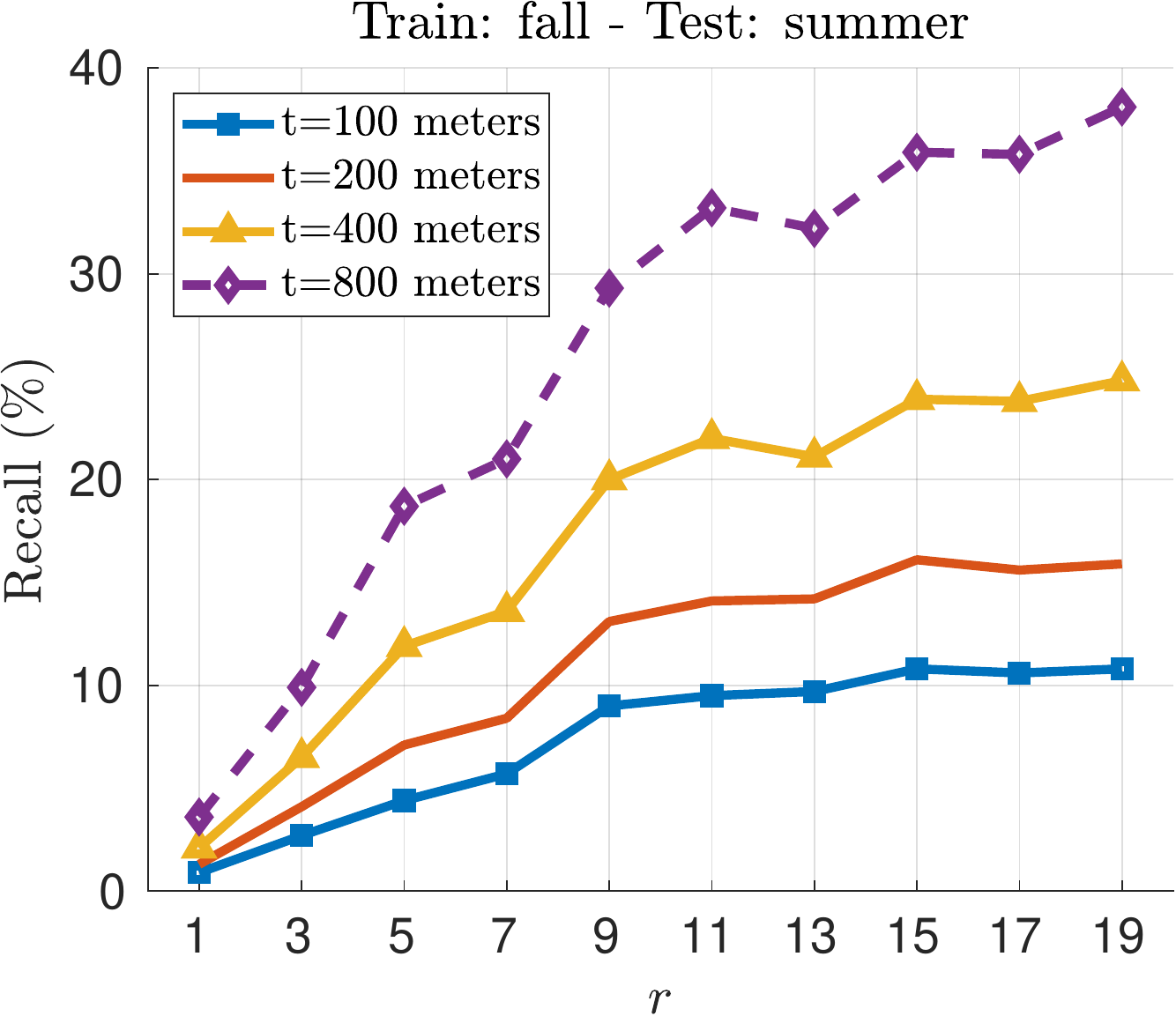}
    \includegraphics[width = 0.49\columnwidth]{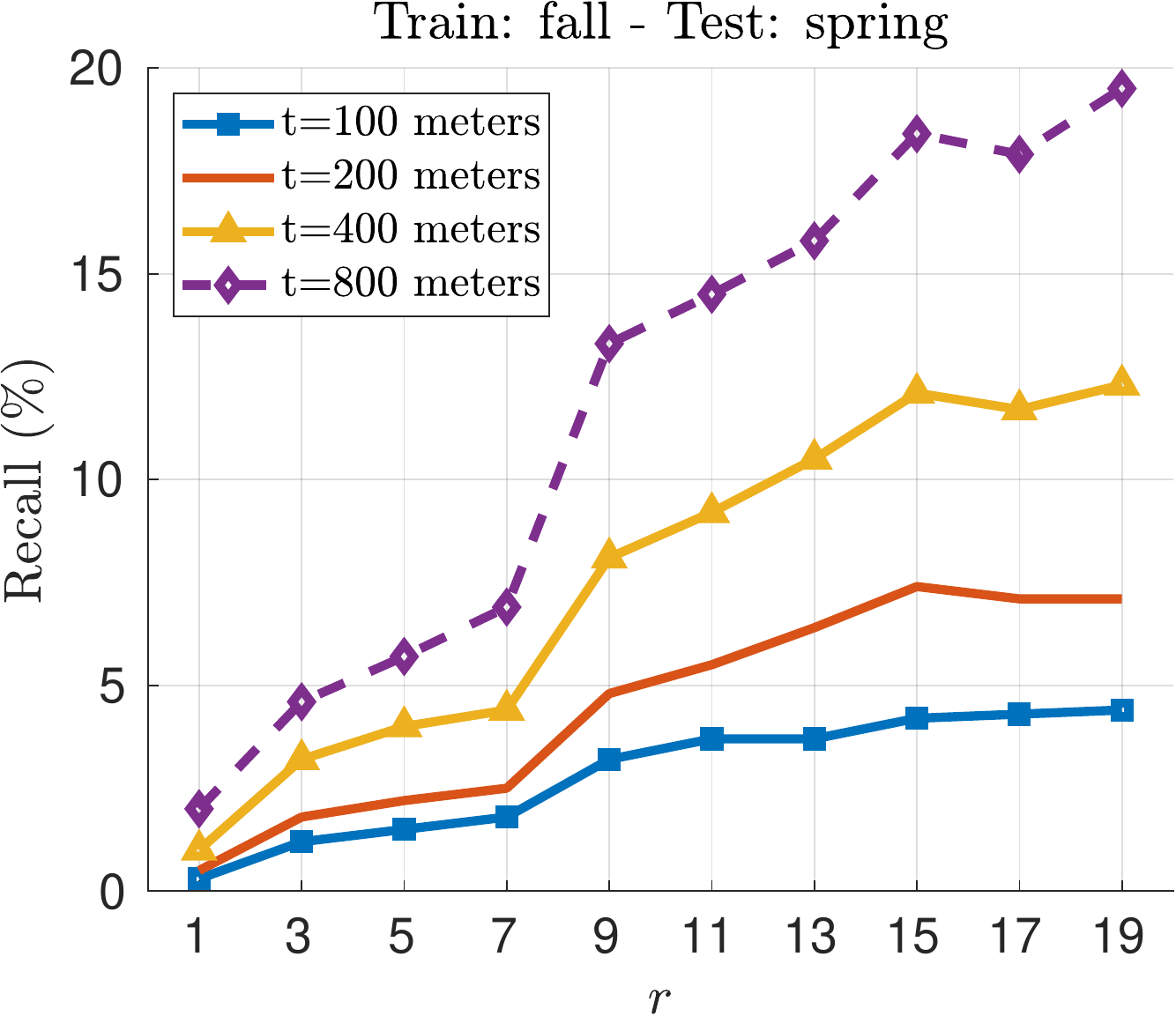}
    \caption{Performance of the compressed training scheme with varying values of $r$ when $d'=0.1d$.}
    \label{fig:region_analyze}
\end{figure}

In the next analysis, we study the system's performance with varying values of $r$. We fix the values of $d' = 0.1d$ and execute the methods when $r$ increases from $1$ to $20$. The results are shown in Fig.~\ref{fig:region_analyze}. \add{Note that when $r$ increases, the system achieves better performance since the trees corresponding to the sub-regions can be better trained with location-specific data.  Fig.~\ref{fig:region_analyze} also shows that BTEL achieves satisfactory performance, even for $r<10$, which demonstrates the usefulness of the underlying tree structure, while the clustering of the training data can be optionally employed to improve the overall performance.}


%% file: tex/conclusion.tex
\section{Conclusion and Future Work}
We have presented a novel binary tree encoding approach for visual localization that achieves both sub-linear storage growth and sub-linear query time with respect to the number of images in the training database. By combining our novel encoding approach with efficient feature selection and sequence summarization, our method outperforms existing state-of-the-art approaches under limited storage constraint. Compared to the recently proposed method~\cite{yu2018rhythmic} for sub-linear storage, our method provides substantially better localization accuracy and requires much less storage footprint. Moreover, our method is more robust to changes in visual conditions. The encoding scheme also allows the memory to be easily configured to suit the application at hand. Also, the fact that our system is agnostic to the input descriptors enables our method to be used in several real-world applications where the environments undergo drastic changes in visual conditions by taking advantage of state-of-the-art global image descriptors. Additionally, the localization results can also be boosted by and optional sequence filtering technique.

Future work can consider the methods for improving the linear classifiers. 
\add{One promising direction is to investigate the integration of per-location calibration as suggested in~\cite{gronat2013learning} to the existing tree structure in order to achieve sub-linear storage growth}. Our proposed algorithm has the potential to be used in other application domains such as voice recognition, face and object recognition, as long as good global descriptors are provided, which can also be further explored in future work.

%% file: main.bbl
\begin{thebibliography}{10}
\providecommand{\url}[1]{#1}
\csname url@samestyle\endcsname
\providecommand{\newblock}{\relax}
\providecommand{\bibinfo}[2]{#2}
\providecommand{\BIBentrySTDinterwordspacing}{\spaceskip=0pt\relax}
\providecommand{\BIBentryALTinterwordstretchfactor}{4}
\providecommand{\BIBentryALTinterwordspacing}{\spaceskip=\fontdimen2\font plus
\BIBentryALTinterwordstretchfactor\fontdimen3\font minus
  \fontdimen4\font\relax}
\providecommand{\BIBforeignlanguage}[2]{{%
\expandafter\ifx\csname l@#1\endcsname\relax
\typeout{** WARNING: IEEEtran.bst: No hyphenation pattern has been}%
\typeout{** loaded for the language `#1'. Using the pattern for}%
\typeout{** the default language instead.}%
\else
\language=\csname l@#1\endcsname
\fi
#2}}
\providecommand{\BIBdecl}{\relax}
\BIBdecl

\bibitem{sattler2018benchmarking}
T.~Sattler, W.~Maddern, C.~Toft, A.~Torii, L.~Hammarstrand, E.~Stenborg,
  D.~Safari, M.~Okutomi, M.~Pollefeys, J.~Sivic \emph{et~al.}, ``Benchmarking
  {6DOF} outdoor visual localization in changing conditions,'' in \emph{CVPR},
  2018.

\bibitem{lowry2016visual}
S.~Lowry, N.~S{\"u}nderhauf, P.~Newman, J.~J. Leonard, D.~Cox, P.~Corke, and
  M.~J. Milford, ``Visual place recognition: A survey,'' \emph{IEEE
  Transactions on Robotics}, vol.~32, no.~1, pp. 1--19, 2016.

\bibitem{milford2012seqslam}
M.~J. Milford and G.~F. Wyeth, ``Seqslam: Visual route-based navigation for
  sunny summer days and stormy winter nights,'' in \emph{Robotics and
  Automation (ICRA), 2012 IEEE International Conference on}.\hskip 1em plus
  0.5em minus 0.4em\relax IEEE, 2012, pp. 1643--1649.

\bibitem{se2002mobile}
S.~Se, D.~Lowe, and J.~Little, ``Mobile robot localization and mapping with
  uncertainty using scale-invariant visual landmarks,'' \emph{The international
  Journal of robotics Research}, vol.~21, no.~8, pp. 735--758, 2002.

\bibitem{cummins2008fab}
M.~Cummins and P.~Newman, ``Fab-map: Probabilistic localization and mapping in
  the space of appearance,'' \emph{The International Journal of Robotics
  Research}, vol.~27, no.~6, pp. 647--665, 2008.

\bibitem{netvlad}
R.~Arandjelovi{\'c}, P.~Gronat, A.~Torii, T.~Pajdla, and J.~Sivic, ``Net{VLAD}:
  {CNN} architecture for weakly supervised place recognition,'' in \emph{CVPR},
  2016.

\bibitem{jegou2011product}
H.~J{\'e}gou, M.~Douze, and C.~Schmid, ``Product quantization for nearest
  neighbor search,'' \emph{IEEE TPAMI}, vol.~33, no.~1, pp. 117--128, 2011.

\bibitem{yu2018rhythmic}
L.~Yu, A.~Jacobson, and M.~Milford, ``Rhythmic representations: Learning
  periodic patterns for scalable place recognition at a sublinear storage
  cost,'' \emph{IEEE Robotics and Automation Letters}, vol.~3, no.~2, pp.
  811--818, 2018.

\bibitem{krajnik2017fremen}
T.~Krajn{\'\i}k, J.~P. Fentanes, J.~M. Santos, and T.~Duckett, ``Fremen:
  Frequency map enhancement for long-term mobile robot autonomy in changing
  environments,'' \emph{IEEE Transactions on Robotics}, vol.~33, no.~4, pp.
  964--977, 2017.

\bibitem{gong2013iterative}
Y.~Gong, S.~Lazebnik, A.~Gordo, and F.~Perronnin, ``Iterative quantization: A
  procrustean approach to learning binary codes for large-scale image
  retrieval,'' \emph{IEEE Transactions on Pattern Analysis and Machine
  Intelligence}, vol.~35, no.~12, pp. 2916--2929, 2013.

\bibitem{nister2006scalable}
D.~Nister and H.~Stewenius, ``Scalable recognition with a vocabulary tree,'' in
  \emph{Computer vision and pattern recognition, 2006 IEEE computer society
  conference on}, vol.~2.\hskip 1em plus 0.5em minus 0.4em\relax Ieee, 2006,
  pp. 2161--2168.

\bibitem{torii201524}
A.~Torii, R.~Arandjelovic, J.~Sivic, M.~Okutomi, and T.~Pajdla, ``24/7 place
  recognition by view synthesis,'' in \emph{Proceedings of the IEEE Conference
  on Computer Vision and Pattern Recognition}, 2015, pp. 1808--1817.

\bibitem{finetune_hard_samples}
F.~Radenovi{\'c}, G.~Tolias, and O.~Chum, ``{CNN} image retrieval learns from
  {B}o{W}: Unsupervised fine-tuning with hard examples,'' in \emph{ECCV}, 2016.

\bibitem{do2018selective}
T.-T. Do, T.~Hoang, D.-K.~L. Tan, H.~Le, and N.-M. Cheung, ``From selective
  deep convolutional features to compact binary representations for image
  retrieval,'' \emph{arXiv preprint arXiv:1802.02899}, 2018.

\bibitem{gersho2012vector}
A.~Gersho and R.~M. Gray, \emph{Vector quantization and signal
  compression}.\hskip 1em plus 0.5em minus 0.4em\relax Springer Science \&
  Business Media, 2012, vol. 159.

\bibitem{ge2013optimized}
T.~Ge, K.~He, Q.~Ke, and J.~Sun, ``Optimized product quantization for
  approximate nearest neighbor search,'' in \emph{Computer Vision and Pattern
  Recognition (CVPR), 2013 IEEE Conference on}.\hskip 1em plus 0.5em minus
  0.4em\relax IEEE, 2013, pp. 2946--2953.

\bibitem{kalantidis2014locally}
Y.~Kalantidis and Y.~Avrithis, ``Locally optimized product quantization for
  approximate nearest neighbor search,'' in \emph{Proceedings of the IEEE
  Conference on Computer Vision and Pattern Recognition}, 2014, pp. 2321--2328.

\bibitem{invertedindex}
H.~J{\'e}gou, R.~Tavenard, M.~Douze, and L.~Amsaleg, ``Searching in one billion
  vectors: re-rank with source coding,'' in \emph{2011 IEEE International
  Conference on Acoustics, Speech and Signal Processing (ICASSP)}.\hskip 1em
  plus 0.5em minus 0.4em\relax IEEE, 2011, pp. 861--864.

\bibitem{invertedmultiindex}
A.~Babenko and V.~Lempitsky, ``The inverted multi-index,'' \emph{IEEE
  transactions on pattern analysis and machine intelligence}, vol.~37, no.~6,
  pp. 1247--1260, 2015.

\bibitem{witten2010framework}
D.~M. Witten and R.~Tibshirani, ``A framework for feature selection in
  clustering,'' \emph{Journal of the American Statistical Association}, vol.
  105, no. 490, pp. 713--726, 2010.

\bibitem{mcmanus2015learning}
C.~McManus, B.~Upcroft, and P.~Newman, ``Learning place-dependant features for
  long-term vision-based localisation,'' \emph{Autonomous Robots}, vol.~39,
  no.~3, pp. 363--387, 2015.

\bibitem{potapov2014category}
\BIBentryALTinterwordspacing
D.~Potapov, M.~Douze, Z.~Harchaoui, and C.~Schmid, ``{Category-specific video
  summarization},'' in \emph{{ECCV 2014 - European Conference on Computer
  Vision}}, 2014. [Online]. Available: \url{http://hal.inria.fr/hal-01022967}
\BIBentrySTDinterwordspacing

\bibitem{gronat2013learning}
P.~Gronat, G.~Obozinski, J.~Sivic, and T.~Pajdla, ``Learning and calibrating
  per-location classifiers for visual place recognition,'' in \emph{Proceedings
  of the IEEE conference on computer vision and pattern recognition}, 2013, pp.
  907--914.

\end{thebibliography}
